%% file: main.tex
\documentclass[11pt]{article}
\usepackage{amsmath, amssymb, amscd, amsthm, amsfonts}
\usepackage{float}
\usepackage{graphicx}
\usepackage[hidelinks]{hyperref}
\usepackage{tabularx}
\usepackage{booktabs}
\usepackage{natbib}
\usepackage{xspace}
\usepackage{adjustbox}
\usepackage[most]{tcolorbox}
\usepackage{authblk}

\title{PublicAgent: Multi-Agent Design Principles From an LLM-Based Open Data Analysis Framework}

\date{September 2025}

\newcommand{\modelname}{PublicAgent\xspace}

\author{Sina Montazeri}
\author{Yunhe Feng}
\author{Kewei Sha}
\affil{University of North Texas}
\affil{\texttt{sina.montazeri@my.unt.edu} \quad \texttt{Yunhe.Feng@unt.edu} \quad \texttt{Kewei.Sha@unt.edu}}

\begin{document}

\maketitle

\input{sections/abstract}

\input{sections/introduction}

\input{sections/related_work}
\input{sections/methodology}
\input{sections/evaluation}
\input{sections/results}
\input{sections/conclusion}

\bibliographystyle{unsrt}
\bibliography{references}

\end{document}

%% file: sections/abstract.tex
\begin{abstract}
    Open data repositories hold potential for evidence-based decision-making, yet remain inaccessible to non-experts who lack expertise in dataset discovery, schema mapping, and statistical analysis. While large language models show promise for individual tasks, end-to-end analytical workflows expose fundamental limitations: attention dilutes across growing contexts, specialized reasoning patterns interfere, and errors propagate without detection. We present \modelname, a multi-agent framework that addresses these limitations through targeted decomposition into specialized agents for intent clarification, dataset discovery, analysis, and reporting. This architecture maintains focused attention within agent contexts and enables validation at each pipeline stage. Evaluation across five models and 50 queries derives five design principles for multi-agent LLM systems. First, specialization provides value independent of model strength—even the strongest model shows 97.5\% agent win rates, demonstrating benefits orthogonal to model scale. Second, agents divide into universal (discovery, analysis) and conditional (report, intent) categories, with universal agents showing consistent effectiveness (std dev 12.4\%) while conditional agents vary by model (std dev 20.5\%). Third, agents mitigate distinct failure modes—removing discovery or analysis causes catastrophic failures (243-280 instances), while removing report or intent causes quality degradation. Fourth, architectural benefits persist across task complexity with stable win rates (86-92\% analysis, 84-94\% discovery), indicating workflow management value rather than reasoning enhancement. Fifth, wide variance in agent effectiveness across models (42-96\% for analysis) requires model-aware architecture design. These principles guide when and why specialization is necessary for complex analytical workflows while enabling broader access to public data through natural language interfaces.
\end{abstract}

%% file: sections/introduction.tex
\section{Introduction}
Open and public data serve as critical resources for improving transparency, supporting innovation, and promoting equitable development. When accessible to a broader audience, these datasets enable communities to tackle pressing issues, refine policies, and support informed discourse. Yet, the sheer volume, intricate structures, and dispersed nature of such data—exemplified by the U.S. Government's data.gov portal with nearly 300,000 datasets—impose heavy demands on users seeking to derive meaningful conclusions \citep{Hilbert2011, Boyd2012}.

Realizing this potential requires overcoming substantial technical barriers. Users without data mining and analysis expertise encounter obstacles at every stage of the analytical workflow. Identifying pertinent datasets demands semantic search across scattered repositories with inconsistent metadata standards. Once located, datasets frequently arrive in heterogeneous formats that require normalization and schema mapping to resolve inconsistencies. Transforming raw data into insights involves statistical computation and code generation, skills inaccessible to journalists, policymakers, and community advocates without technical training. These compounding barriers restrict open data utilization to a narrow technical audience, undermining the democratic potential of public information.

To address these accessibility challenges, large language models (LLMs) emerge as a promising solution with demonstrated capabilities in information retrieval, data manipulation, and code synthesis through natural language interfaces \citep{chen2023beyond, faggioli2023perspectives, li2023synthetic, lee2023making, li2024large, muffoletto2024l2ceval}. Prior efforts have applied LLMs to individual stages of data workflows: generating retrieval-augmenting knowledge, assessing dataset relevance, creating synthetic data, and evaluating code generation. However, these approaches address isolated components rather than the full analytical pipeline from ambiguous query to evidence-based report.

Applying LLMs to the complete analytical workflow reveals fundamental limitations of single-model approaches. A complete workflow requires semantic search across repositories, schema mapping to diverse formats, statistical analysis with validation, and report synthesis with appropriate caveats—each demanding specialized reasoning patterns. When a single LLM handles all stages sequentially, attention dilutes across the growing context window, specialized skills interfere (semantic search reasoning differs from statistical analysis), and errors propagate without detection. Consider a journalist asking ``How prevalent is adult high blood pressure in the United States?'' The system must clarify ``adult'' (age 18+?) and ``high blood pressure'' (clinical threshold?), search heterogeneous health repositories, map to correct schema fields, execute validated statistical code, and present results with dataset provenance and limitations. Asking one model to maintain coherent reasoning across these distinct competencies in a single long-context invocation leads to semantic drift, factual errors, and incomplete analysis.

To overcome these fundamental limitations, we present \modelname, a multi-agent framework that decomposes the analytical pipeline into specialized agents with distinct responsibilities: intent clarification resolves query ambiguities, data discovery performs semantic search and metadata synthesis, data analysis generates and validates statistical code, and report generation synthesizes findings with appropriate caveats. This specialization prevents task interference, enables targeted validation at each stage, and maintains focused attention within agent contexts. Natural language interfaces at every stage eliminate the need for SQL, Python, or API knowledge, democratizing access to users without technical expertise.

Our main contributions include:
\begin{itemize}
\item Five design principles for multi-agent LLM systems that establish when and why specialization is necessary, covering: (1) value independent of model strength, (2) universal versus conditional agent deployment, (3) distinct failure mode mitigation, (4) complexity-independent architectural benefits, and (5) model-aware architecture design.
\item Design and implementation of \modelname, a multi-agent framework for open data analysis that serves as the empirical foundation for deriving these principles through systematic ablation studies across specialized agents for intent clarification, dataset discovery, analysis, and reporting.
\item Empirical validation across five language models and 50 diverse queries demonstrating that these principles hold independent of base model capability, with comprehensive ablations isolating each agent's contribution and revealing how specialization addresses attention dilution, task interference, and error propagation.
\item Characterization of fundamental barriers in end-to-end open data analysis—dataset discovery across heterogeneous repositories, schema mapping, statistical validation—that necessitate multi-agent approaches beyond what single models can achieve, while enabling broader access through natural language interfaces.
\end{itemize}

%% file: sections/related_work.tex
\section{Related Work}
Natural language interfaces for data analysis represent a convergence of multiple research domains. Prior work has made substantial progress on individual components—semantic parsing, automated analysis, multi-agent coordination, query refinement, and report generation—but no integrated system addresses the complete pipeline from ambiguous natural language queries to comprehensive reports using discovered open datasets. We organize the related work into five key areas, identifying what each contributes and what gaps remain.

\subsection{Natural Language Interfaces to Databases}
Early work on natural language interfaces to databases (NLIDB) focused on translating user queries into structured database queries. LUNAR demonstrated rule-based semantic parsing for database queries~\citep{woods1972lunar}. This foundation allowed for more sophisticated neural approaches in the subsequent years. Text-to-SQL systems like SQLNet~\citep{xu2017sqlnet} and RAT-SQL~\citep{wang2020ratsql} use sequence-to-sequence models to map questions to SQL queries. These systems achieve substantial accuracy on benchmarks like Spider~\citep{yu2018spider} and WikiSQL~\citep{zhong2017seq2sql}, although they typically assume a known database schema and focus on single-step query generation.

The integration of large language models (LLMs) has further advanced semantic parsing capabilities. BIRD~\citep{li2024bird} and DAIL-SQL~\citep{gao2023dail} leverage pre-trained models with few-shot prompting to handle complex queries across diverse schemas. These advances represent significant progress in query translation for known databases, yet remain limited to pre-existing schemas. Open data analysis requires additional capabilities: discovering relevant datasets across heterogeneous repositories and synthesizing metadata when publisher information is incomplete or inconsistent.

\subsection{Automated Data Analysis and AutoML}
Although semantic parsing assumes data access, automated analysis systems address the complementary challenge of extracting insights once datasets are available.
Automated machine learning (AutoML) frameworks have evolved to address increasingly complex data science workflows. Early systems like Auto-WEKA~\citep{kotthoff2017auto}, AutoSklearn~\citep{feurer2015efficient} and TPOT~\citep{olson2016tpot} automate model selection and hyperparameter tuning for predictive tasks. More recent systems extend this automation to broader aspects of data analysis. AIDE~\citep{kaddour2023aide} generates exploratory data analysis code, while Data Formulator~\citep{wang2024data} enables iterative chart specification through natural language. These systems excel at specific analytical tasks, but require users to provide datasets. They often focus on visualization or model training rather than comprehensive reporting.

Systems such as LIDA~\citep{dibia2023lida} and ChartLLM~\citep{han2023chartllm} move closer to end-to-end generation. LIDA generates visualizations from data through goal-driven exploration. ChartLLM produces chart summaries using LLMs. Both systems demonstrate the potential of automated insight generation, yet they assume datasets are pre-loaded and focus on visualization rather than comprehensive analytical workflows spanning query clarification, dataset discovery, multi-faceted analysis, and narrative synthesis.

\subsection{Multi-Agent Systems for Complex Tasks}
Recognizing that isolated capabilities are insufficient for complex workflows, multi-agent architectures have emerged as effective strategies for decomposing tasks into specialized subtasks. This paradigm recognizes that diverse capabilities are better handled through specialization than monolithic approaches. MetaGPT~\citep{hong2023metagpt} uses software development roles (architect, engineer, tester) to generate code collaboratively. AutoGen~\citep{wu2023autogen} provides a framework for conversational agents that coordinate through the transmission of messages. TaskWeaver~\citep{qiao2024taskweaver} decomposes data analysis tasks into code snippets executed by specialized agents. These systems demonstrate performance improvements when tasks require multiple competencies.

The success of multi-agent collaboration extends to data science and information retrieval. Co-STORM~\citep{shao2024costorm} demonstrates multi-agent collaboration for information seeking. ChatDev~\citep{qian2023chatdev} applies collaborative agents in software engineering. However, these systems primarily address code generation or web search within controlled environments where data sources, formats, and quality are known. Open data repositories present distinct challenges: semantic search across inconsistent metadata, format normalization without predefined schemas, and validation across varying data quality levels.

\textcolor{blue}{The orchestration of multi-agent systems varies in how control and coordination are implemented. HuggingGPT~\citep{shen2023hugginggpt} positions an LLM as a central controller managing a four-stage pipeline—task planning, model selection, execution, and response generation—demonstrating sequential orchestration for multi-modal AI tasks. MegaAgent~\citep{wang2025megaagent} extends this to large-scale coordination through dynamic task decomposition and parallel execution, minimizing human-designed prompts by having agents autonomously generate coordination procedures. AgentVerse~\citep{chen2023agentverse} explores collaborative orchestration where expert agents are assembled into automatic systems exhibiting emergent behaviors in reasoning and tool utilization. Recent analyses~\citep{li2024agentplanning, tran2025multiagent, han2024llmmas} categorize orchestration approaches along dimensions of centralization (centralized coordinators versus peer-to-peer), temporal structure (interleaved versus decomposition-first), and coordination protocols, finding that explicit orchestration mechanisms reduce error propagation and attention dilution in complex workflows.}

\subsection{Query Understanding and Disambiguation}
Even with multi-agent coordination and automated analysis capabilities, the full pipeline requires addressing ambiguities present in initial natural language queries.
Effective data analysis begins with a precise query formulation. Ambiguities in initial queries can propagate through the analytical pipeline and compromise results. Research on query clarification addresses these issues through interactive dialogs. Elicitron~\citep{zamani2020generating} learns to ask clarifying questions for information retrieval tasks. ClariQ~\citep{aliannejadi2020convai3} provides a benchmark for conversational query refinement. These systems improve retrieval accuracy through iterative refinement. In database contexts, systems like NaLIR~\citep{li2014constructing} identify ambiguous entities and relationships. They prompt users to resolve ambiguities before the execution of the query.

Large language models have enabled new approaches to query enhancement. Query2Doc~\citep{wang2023query2doc} generates pseudo-documents to enhance retrieval. CoT-BERT~\citep{wu2022cot} decomposes complex queries into reasoning chains. These approaches primarily target general search or retrieval tasks. Data analysis requires resolving domain-specific ambiguities—undefined age thresholds, imprecise medical terminology, and unclear temporal scopes—that directly affect dataset discovery and statistical computation accuracy.

\subsection{Data-to-Text and Report Generation}
Finally, once data is discovered, analyzed, and validated, results must be synthesized into accessible reports for non-expert audiences. Automated report generation from structured data has progressed substantially from the early template-based approaches. Early data-to-text systems like FoG~\citep{goldberg1994using} and SumTime~\citep{reiter2005choosing} relied on hand-made rules. The introduction of neural models such as LSTM encoders and Transformers enabled more flexible generation. Recent work on table-to-text generation includes TAPAS~\citep{herzig2020tapas} for question answering over tables and GPT-3-based approaches for numerical reasoning~\citep{chen2022finqa}. These advances have improved fluency and coherence in generated text.

Applications in data analysis contexts have demonstrated practical value. Quill~\citep{allen2010stats} generates narrative insights from business intelligence data. Narrativa~\citep{gatt2018survey} produces automated journalism from structured sources. These systems deliver compelling narratives, but assume clean, well-structured inputs and focus on linguistic fluency. Multi-stage analytical pipelines require maintaining traceability across components—query clarification, dataset provenance, analytical methods, validation results—while adapting technical content for non-expert audiences.

The preceding review reveals substantial progress on individual components—semantic parsing for known databases, automated analysis for pre-loaded datasets, multi-agent coordination for controlled environments, and report generation for clean inputs. However, no system integrates these capabilities for heterogeneous open data repositories where datasets must be discovered across inconsistent metadata, schemas are unknown, and quality varies. Our framework addresses this gap through a multi-agent architecture designed specifically for open data challenges: intent clarification targeting analytical ambiguities, dataset discovery with metadata synthesis, validated statistical analysis, and traceable report generation. Through systematic ablations across five models, we demonstrate when and why agent specialization provides value, characterizing the failure modes—attention dilution, task interference, error propagation—that necessitate decomposition for complex analytical workflows.

%% file: sections/methodology.tex
\section{Methodology}

Transforming natural language queries into evidence-based analytical reports requires bridging semantic ambiguity, heterogeneous data formats, statistical computation, and narrative synthesis. We decompose this task into specialized functions implemented by dedicated agents: intent clarification resolves query ambiguities, data discovery performs semantic search and metadata synthesis across repositories, data analysis generates and validates statistical code, and report generation synthesizes findings with appropriate limitations. This section formalizes the problem, describes each agent's implementation, and details the coordination mechanisms that maintain coherence across the pipeline.

\subsection{Problem Formulation}

We formalize the task of converting a user's natural language query $Q^u$ into a comprehensive report $R$ derived from a collection of open and public datasets $D = \{D_1, \dots, D_n\}$. This process addresses the challenge of mapping unstructured queries to structured data across heterogeneous repositories. The overall solution is an orchestration function $f_o$ that coordinates specialized agents:
\begin{align}
R = f_o(Q^u, D) \text{ where } f_o \text{ coordinates } (f_q, f_d, f_x, f_g).
\end{align}

The orchestration function $f_o$ manages specialized functions that handle query refinement, dataset discovery, data analysis, and report generation: $f_q$, $f_d$, $f_x$, and $f_g$, respectively.

These functions operate sequentially under orchestrator coordination to respect information dependencies: query refinement must precede dataset discovery to avoid propagating ambiguities, dataset discovery must inform analysis to ensure schema compatibility, and report generation requires all prior outputs for synthesis.

The query refinement function $f_q$ improves the original query $Q^u$ through clarification:
\begin{align}
Q^e = f_q(Q^u).
\label{equ:f_q}
\end{align}

The dataset discovery function $f_d$ selects a relevant dataset $D_i \in D$ and enriches it with comprehensive metadata $M$ synthesized from multiple sources: publisher-provided schemas, computed statistical summaries, structural information (column names, types, row counts), sample data excerpts, and dataset provenance details:
\begin{align}
(D_i, M) = f_d(Q^e, D).
\label{equ:f_d}
\end{align}

The data analysis function $f_x$ takes as input the enhanced query, selected dataset, and its metadata:

\begin{align}
E = f_x(Q^e, D_i, M),
\end{align}

where $E$ represents a collection of analytical experiments, each implemented as an executable Python program that operates on $D_i$ to extract insights aligned with $Q^e$. The function dynamically generates experiments based on query requirements—from simple aggregations to complex multi-step analyses involving filtering, grouping, and statistical computations.

Finally, the report generation function $f_g$ synthesizes these elements into the final report $R$:
\begin{align}
R = f_g(Q^u, Q^e, D_i, E).
\label{equ:f_g}
\end{align}

To evaluate the report, we define a quality function $Q(R)$ assessing its performance across multiple dimensions:
\[
Q(R) = \tfrac{1}{4}(F(R) + C(R) + V(R) + H(R)),
\]
where $F(R)$, $C(R)$, $V(R)$, and $H(R)$ are scores (1-10) for factual consistency, completeness, relevance, and coherence, respectively, derived from rubric-based judgments. Each component of the pipeline thus contributes to producing a report $R$ that maximizes $Q(R)$ for a given query $q$.

Each component thus contributes to a coherent workflow from query to report.

\subsection{Framework Overview}
Building on the problem formulation, \modelname implements $f_o$ through a multi-agent pipeline that distributes responsibilities across specialized agents, each optimized for distinct aspects of the task. This decomposition exploits domain-specific processing strengths over monolithic agents, which falter when handling varied competencies such as linguistic refinement, data retrieval, statistical computation, and narrative synthesis.

\modelname proceeds sequentially to respect information dependencies: query clarification must precede dataset discovery to avoid propagating ambiguities; dataset discovery must inform analysis to ensure schema compatibility; and report generation requires all prior outputs for synthesis. Semantic consistency is maintained via structured message passing and summarization, where each agent validates its inputs and outputs against workflow invariants such as query intent and dataset provenance.

\begin{figure}[htbp]
    \centering
    \includegraphics[width=1\linewidth]{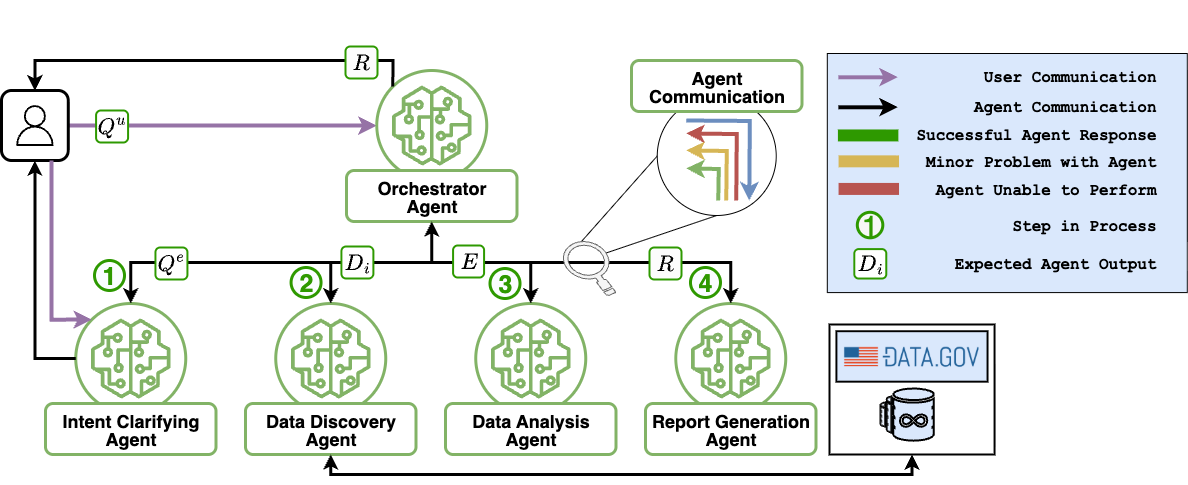}
    \caption{Overall Structure}
    \label{fig:overall_framewor}
\end{figure}

Figure~\ref{fig:overall_framewor} illustrates this structure, with agents for intent clarification ($A_{IC}$), data discovery ($A_{DD}$), data analysis ($A_{DA}$), and report generation ($A_{RG}$) operating under an orchestrator. This setup preserves fidelity from $Q^u$ to $R$, aligning with the functions defined in the problem formulation.

\subsection{Orchestrator Agent}
The orchestrator implements the coordination for $f_o$ through sequential invocation of specialized agents, ensuring diverse capabilities required for the workflow are applied while preserving coherence. It organizes the process into stages aligned with $f_q$, $f_d$, $f_x$, and $f_g$ and validates progress at each stage through evaluations of outputs such as refined queries, selected datasets, analysis results and generated reports.

The orchestrator passes context between stages and maintains information like original and enhanced queries, dataset details, and analysis outcomes. In the event of inconsistencies or errors, it identifies issues and initiates retries or adjustments, categorizing problems to apply appropriate resolutions: temporary failures lead to re-attempts, while persistent mismatches prompt re-evaluation of prior stages.

This approach allows reliable progression from $Q^u$ to $R$, integrating the formalized functions into an operational sequence that adapts to potential disruptions in the multi-agent interaction.

\subsection{Intent Clarifying Agent}

Natural language queries for data analysis often contain colloquial terms, ambiguous scopes, and implicit assumptions that prevent effective dataset retrieval and analysis, as terms like ``adults,'' ``high blood pressure,'' or ``recent'' lack the precision necessary for matching dataset schemas or formulating analytical operations. Prior work either ignores this ambiguity (leading to semantic mismatches) or requires users to learn structured query languages (creating accessibility barriers), so the challenge is automatically transforming informal language into scientifically precise specifications while preserving user intent and conversational accessibility.

The intent clarifying agent implements $f_q$ by identifying and resolving ambiguities in $Q^u$ to produce an enhanced query $Q^e$. The agent detects up to three critical ambiguities—imprecise terms, unclear temporal or geographic scopes, and undefined thresholds—and proposes precise substitutions framed as confirmation questions. This iterative process continues until all identified issues are resolved.

Consider a user query $Q^u$: ``How many adults have high blood pressure in the US?'' The agent detects ambiguities in ``adults'' and ``high blood pressure,'' proposing to replace ``adults'' with ``individuals aged 18 and older'' and ``high blood pressure'' with ``diagnosed hypertension (blood pressure $\geq$ 140/90 mmHg)''. After confirmation, the enhanced query $Q^e$ becomes:

\[
\text{How many } \underbrace{\text{adults}}_{\text{individuals aged 18 and older}} \text{ have } \underbrace{\text{high blood pressure}}_{\text{diagnosed hypertension (BP $\geq$ 140/90 mmHg)}} \text{ in the US?}
\]

This approach bridges colloquial language and analytical precision without requiring domain expertise, ensuring $Q^e$ retains the intent of $Q^u$ while providing specificity for dataset discovery and analysis.

\subsection{Data Discovery Agent}

Open data repositories expose datasets through heterogeneous APIs with inconsistent metadata standards, varying response formats, and unpredictable quality, requiring precise queries like ``diagnosed hypertension prevalence in individuals aged 18+'' to be mapped to relevant datasets across repositories with different terminologies, granularities, and access patterns. Existing approaches either rely on manual dataset curation (not scalable) or simple keyword matching (semantically inadequate), so the challenge is automated semantic search that handles vocabulary mismatches, evaluates dataset relevance, adapts to format inconsistencies, and synthesizes comprehensive metadata from fragmented sources.

The data discovery agent implements $f_d$ to retrieve and prepare a relevant dataset $D_i$ with enriched metadata $M$. It generates search terms from $Q^e$ and queries repositories such as data.gov, evaluating results for semantic alignment with query intent. When multiple candidates exist, the agent selects the most suitable by domain relevance and schema compatibility; when none match, it adaptively broadens search parameters (e.g., expanding geographic scope from county to state level). The agent prioritizes CSV formats for compatibility and employs fallback mechanisms to extract direct download links when API responses are non-standard.

Metadata synthesis integrates publisher-provided information with insights from dataset inspection, and Table~\ref{tab:metadata_content} outlines the standard components of metadata $M$ generated by the agent.

\begin{table}[htbp]
\caption{Components of Standard Metadata $M$ Generated by the Data Discovery Agent}
\label{tab:metadata_content}
\centering
\begin{tabularx}{\textwidth}{lX}
\toprule
\textbf{Component} & \textbf{Description} \\
\midrule
First 5 rows & Initial rows offering a snapshot of dataset content. \\
Column names & List of all column headers in the dataset. \\
Number of rows & Total count of data entries. \\
Head output & Summary of the first five rows, including values for all columns. \\
Tail output & Summary of the last five rows, including values for all columns. \\
Statistical description & Descriptive statistics (e.g., mean, median, standard deviation) for numerical columns. \\
Unique column values & Up to ten unique values per column to show data diversity. \\
Publisher metadata & Publisher-provided codebook with column descriptions and data types. \\
\bottomrule
\end{tabularx}
\end{table}

Error handling distinguishes recoverable issues (transient download failures) from fundamental mismatches (semantic incompatibility with $Q^e$), ensuring reliable alignment with the query and preparing the dataset for subsequent analysis.

\subsection{Data Analysis Agent}

Translating a natural language query into executable analytical code requires bridging semantic intent with computational operations, complicated by heterogeneous dataset schemas, diverse analytical requirements, and validation needs. A query like ``diagnosed hypertension prevalence by age subgroups'' must be decomposed into specific computations (filtering, aggregation, subgroup analysis), mapped to correct dataset columns, implemented as executable Python code, and validated for correctness. Existing solutions either require manual coding (inaccessible to non-experts) or apply rigid templates (inflexible for diverse queries), so the challenge is automated generation of multi-step analytical workflows that adapt to varied schemas, validate intermediate outputs, and handle data quality issues.

The data analysis agent implements $f_x$ to transform the enhanced query $Q^e$, dataset $D_i$, and metadata $M$ into verified insights through a structured analytical pipeline. The agent inspects $M$ (column names, types, statistical summaries) to map query requirements to dataset structure, then decomposes $Q^e$ into discrete experiments representing specific computations.

Consider the enhanced query ``How many individuals aged 18 and over have diagnosed hypertension (blood pressure $\geq 140/90$ mmHg) in the US?'' The agent formulates experiments such as:
\begin{itemize}
    \item Compute the proportion of diagnosed hypertension cases: 
      \[
      P_{\text{hypertension}} = \frac{|\{ x \in D_i \mid x_{\text{hypertension}} = 1 \}|}{|D_i|}
      \]
    \item Calculate the average prevalence by age subgroups (e.g., 18--44, 45--64, 65+):
      \[
      P_{\text{subgroup}} = \frac{|\{ x \in D_i \mid x_{\text{hypertension}} = 1 \land x_{\text{age}} \in S \}|}{|\{ x \in D_i \mid x_{\text{age}} \in S \}|}, \quad S \in \{[18, 44], [45, 64], [65, \infty\}
      \]
\end{itemize}

The agent translates these mathematical formulations into executable code operating on the dataset. Figure~\ref{code:prevalence} demonstrates how the overall prevalence computation is implemented:

\begin{figure}[htbp]
\centering
\begin{verbatim}
# compute overall hypertension prevalence
adults = df[df['age'] >= 18]
hypertension_cases = adults[adults['hypertension'] == 1]
overall_prevalence = len(hypertension_cases) / len(adults)
\end{verbatim}
\caption{Example Python code generated by the data analysis agent for computing hypertension prevalence}
\label{code:prevalence}
\end{figure}
These experiments are executed as self-contained Python programs on a preloaded dataframe with pandas preimported, ensuring no modifications or variables persist between runs to prevent unintended side effects.

The results of each experiment are captured for the specified variables and evaluated for reasonableness. Anomalies, such as zero values indicating potential filtering errors, trigger diagnostic checks. If parsing issues arise (e.g., mixed separators or encoding errors), the agent adapts by testing alternative configurations to ensure reliable handling of heterogeneous data formats. Table~\ref{tab:experiment_outputs} illustrates the sample results of these experiments.

\begin{table}[htbp]
\caption{Sample Experiment Outputs for Hypertension Prevalence Query}
\label{tab:experiment_outputs}
\centering
\begin{tabularx}{\textwidth}{lX}
\toprule
\textbf{Experiment} & \textbf{Output} \\
\midrule
Proportion of hypertension cases & $ P_{\text{hypertension}} = 0.32 $ (32\% of adults) \\
Prevalence, ages 18--44 & $ P_{\text{subgroup}} = 0.15 $ (15\% of subgroup) \\
Prevalence, ages 45--64 & $ P_{\text{subgroup}} = 0.38 $ (38\% of subgroup) \\
Prevalence, ages 65+ & $ P_{\text{subgroup}} = 0.54 $ (54\% of subgroup) \\
\bottomrule
\end{tabularx}
\end{table}

The process iterates until all experiments are completed and validated, aggregating findings traceable to their hypotheses, producing results semantically aligned with $ Q^e $, statistically reliable, and prepared for report generation.

\subsection{Report Generation Agent}

Synthesizing analytical results into accessible reports requires balancing technical precision with readability for diverse audiences, complicated by integrating information from multiple pipeline stages. A report must incorporate the original query context, clarified specifications, dataset provenance, analytical methods, results, limitations, and conclusions while adapting language complexity for non-expert readers without sacrificing accuracy. Existing systems either produce technical outputs inaccessible to general audiences or oversimplified summaries that lack rigor, so the challenge is automated report generation that maintains traceability across workflow stages, adapts terminology to audience expertise, and structures information for coherent narrative flow.

The report generation agent implements $f_{rg}$ to synthesize inputs from previous stages into a structured report $R$ balancing accessibility and precision. It integrates the original query $Q^u$, enhanced query $Q^e$, dataset information, and analysis results into distinct sections outlined in Table~\ref{tab:report_sections}.

\begin{table}[htbp]
\caption{Sections of the Report Generated by the Report Generation Agent}
\label{tab:report_sections}
\centering
\begin{tabularx}{\textwidth}{lX}
\toprule
\textbf{Section} & \textbf{Description} \\
\midrule
Title & A concise heading summarizing the analysis focus. \\
Summary for Non-Experts & A 1-2 paragraph overview explaining main findings in simple language, assuming high school-level knowledge. \\
Analysis Assumptions & Main assumptions underlying the analytical approach, detailing conditions presumed during processing. \\
Analysis Definitions & Explanations of technical terms and concepts used in the analysis, adapted for accessibility. \\
Experiments & Detailed account of conducted experiments, including Experiment Description, Narrative of each experiment's purpose and methodology. Columns Used, Specific dataset columns utilized in each experiment.\\
Limitations & A 1-2 paragraph discussion of potential uncertainties, biases, or constraints in the analysis. \\
Conclusions & A 1-2 paragraph synthesis of main insights derived from the experiments. \\
Dataset Link & Reference to source dataset for traceability and reproducibility. \\
\bottomrule
\end{tabularx}
\end{table}

The agent employs adaptive language modeling to simplify technical terminology for non-expert audiences while preserving analytical precision. It validates that each section is substantive, logically connected, and consistent with prior workflow outputs to prevent information loss during synthesis. Maintaining explicit traceability from $Q^u$ through $Q^e$, dataset selection, analytical experiments, to final conclusions enables verification and reproducibility while remaining accessible to users without technical expertise.

\subsection{Agent Tools}

While large language models possess strong general reasoning capabilities, multi-agent systems require specialized tools to address coordination challenges emerging from task decomposition. When a complex analytical workflow is divided among specialized agents, three fundamental problems arise: (1) agents must maintain coherent state across sequential steps without losing critical information, (2) agents must validate intermediate outputs to prevent error propagation, and (3) agents must interface with external systems (datasets, execution environments) that standard language models cannot directly access.

\modelname provides four tool categories—task management, reasoning augmentation, execution isolation, and data integration—purpose-built for the coordination problems arising when decomposing data analysis into specialized agent responsibilities. Each tool category supports agents differently:

\begin{itemize}
    \item \textbf{Task Management:} All agents use task management to coordinate workflows. The intent clarifying agent tracks query refinement steps, the data discovery agent directs dataset search and retrieval, the data analysis agent orchestrates experiment planning and validation, and the report generation agent tracks section creation.
    
    \item \textbf{Reasoning and Validation:} Each agent applies reasoning tools to its domain. The intent clarifying agent resolves query ambiguities, the data discovery agent validates dataset relevance, the data analysis agent checks experiment outcomes, and the report generation agent confirms report coherence.
    
    \item \textbf{Execution Environment:} Only the data analysis agent requires code execution, using an isolated environment to run analytical experiments on datasets.
    
    \item \textbf{Data Integration:} Three agents interact with external data. The data discovery agent queries repositories and synthesizes metadata, the data analysis agent processes diverse dataset formats, and the report generation agent incorporates dataset details into the final output.
\end{itemize}

\subsubsection{Task Management System}

Language models excel at generating responses to prompts but lack native state management for multi-step processes. When an analysis requires multiple dependent operations—metadata inspection, experiment formulation, execution, validation—agents need explicit mechanisms to track progress, manage dependencies, and coordinate retries. Without structured task management, agents either attempt all steps in a single invocation (overwhelming context and causing errors) or lose track of remaining work.

The task management system maintains a dynamic task list with unique identifiers, status tracking (pending, in-progress, completed), and dependency enforcement. Agents query the system to determine remaining work, mark tasks complete when finished, and spawn new tasks when dependencies are satisfied. This transforms stateless language model invocations into stateful workflow execution. Externalizing state management allows agents to focus on executing individual tasks rather than tracking the entire workflow. Dependency enforcement prevents agents from attempting operations before prerequisites are met (e.g., running analysis before dataset retrieval), while error recovery through task re-queueing enables graceful handling of transient failures without manual intervention. \textcolor{blue}{The system scales to complex queries by decomposing them into manageable subtasks: queries requiring multiple dataset formats, geographic aggregations, or temporal comparisons routinely generate 15-20 tasks with intricate dependency chains, which the system executes reliably through incremental progression and isolated failure recovery.}

Consider the enhanced query ``How many individuals aged 18 and over have diagnosed hypertension (blood pressure $\geq 140/90$ mmHg) in the United States?'' The system generates and tracks tasks for the data analysis stage as shown in Table~\ref{tab:task_example}.

\begin{table}[htbp]
\centering
\caption{Example Task Management for Hypertension Analysis Query}
\label{tab:task_example}
\small
\begin{tabularx}{\textwidth}{clX}
\toprule
ID & Status & Task Description \\
\midrule
1 & Completed & Analyze dataset metadata (columns, types, statistics) \\
2 & Pending & Formulate experiment for overall hypertension proportion \\
3 & Pending & Formulate experiment for age-subgroup prevalence (18--44, 45--64, 65+) \\
4 & Pending & Execute overall proportion, validate for anomalies \\
5 & Pending & Execute subgroup prevalence, validate for reasonableness \\
6 & Pending & Diagnose parsing issues if format inconsistencies occur \\
\bottomrule
\end{tabularx}
\end{table}

Each task has a unique identifier and status tracking. If validation fails, the system re-queues the task with adjusted parameters to ensure reliable progression.

\subsubsection{Thinking and Quality Check Tools}

Multi-agent workflows accumulate context as information flows between agents—original query, refined query, dataset metadata, experiment results. This growing context window causes attention dilution in the transformer attention mechanism. \textcolor{blue}{At each position in the sequence, the model computes how much to "attend to" every other position when processing information. The attention weight determines what fraction of each context position contributes to the current computation.} For a query position $i$ attending to key position $j$ in a sequence of length $N$, \textcolor{blue}{where $q_i$ and $k_j$ are learned representations and $d$ is the embedding dimension,} the attention weight is computed as:
\[
\alpha_{ij} = \frac{\exp(q_i^\top k_j / \sqrt{d})}{\sum_{\ell=1}^N \exp(q_i^\top k_\ell / \sqrt{d})}.
\]
As $N$ grows, the normalizing denominator accumulates more terms, \textcolor{blue}{forcing each individual attention weight $\alpha_{ij}$ to become smaller since all weights must sum to 1. This spreads attention across more positions,} diluting focus on relevant information \citep{Liu2023LostIT, hsieh2024ruler}. Additionally, intermediate outputs such as computed statistics may contain errors that propagate if not detected early.

The thinking tool prompts agents to explicitly articulate their current observation, planned action, and reasoning before proceeding, forcing distillation of relevant context. The quality check tool complements this by applying domain-specific validation rules to intermediate outputs, flagging zero-valued aggregations suggesting data filtering errors and checking for statistical plausibility. Explicit thinking breaks the reasoning process into observable steps, making errors detectable and reducing reliance on implicit context tracking. Quality checks catch errors before they propagate to downstream agents, preventing cascading failures. Together, these tools compensate for the attention dilution and error accumulation inherent in long-context, multi-step workflows.

\subsubsection{Analytical Execution Environment}

Language models cannot directly execute code or access computational environments, yet data analysis requires running Python scripts on actual datasets. The execution environment provides the data analysis agent with isolated execution capability. It maintains a preloaded dataframe with necessary libraries while enforcing strict constraints to prevent persistent changes or unauthorized imports. Each experiment thus runs independently, avoiding unintended interactions between sequential analyses. The environment captures results for specified variables and applies validation checks to confirm computational accuracy. Providing controlled code execution with safety guarantees transforms language model suggestions into verified analytical results.

\subsubsection{Online Open Data Integration}

Open data repositories expose datasets through heterogeneous APIs with varying formats, metadata standards, and access patterns. Language models cannot directly query external services or parse diverse response formats. The data integration component manages dataset queries to repositories such as data.gov, validating format consistency, and synthesizing metadata from multiple sources. It caches previously retrieved datasets to improve performance and adapts to format variations by extracting direct download links from non-standard responses. This combination delivers datasets in consistent formats ready for analysis, abstracting away the complexity of heterogeneous data sources.

%% file: sections/evaluation.tex
\section{Evaluation}
We evaluate \modelname on two axes: report quality for real public-data questions and ablations testing whether specialized agents are needed compared to a single-orchestrator baseline. This setup measures both performance on open-data queries and the contribution of each component.

\subsection{Benchmark Design: Queries, Coverage, and Justification}
\label{sec:benchmark-design}
The benchmark includes $|\mathcal{Q}|=50$ queries reflecting real public-data use cases. Each query represents how non-experts interact with open data: clarifying intent, finding datasets, analyzing tables, and presenting results with sources and caveats. Queries span domains and complexity levels instead of focusing on a single task such as text-to-SQL. They cover health, environment, transportation, campaign finance, public health, and COVID-19 (Table~\ref{tab:benchmark_domains}), matching areas common across open data portals (e.g., Data.gov, NYC Open Data). Each query links to at least one dataset with a clear schema and documented provenance.

We selected datasets using portal metadata such as topic tags, geography, and time coverage from sources like Data.gov. Candidate datasets passed filters for tabular format, adequate coverage, and open licenses. \textcolor{blue}{Existing benchmarks for text-to-SQL (Spider~\citep{yu2018spider}, WikiSQL~\citep{zhong2017seq2sql}, BIRD~\citep{li2023bird}) explicitly avoid ambiguous queries—Spider annotators were instructed to exclude questions that are ``vague or too ambiguous''~\citep{yu2018spider}—and provide known database schemas, resulting in scientifically formulated questions with 40-50\% column-name overlap. While benchmarks like KaggleDBQA~\citep{lee2021kaggledbqa} introduced more natural phrasing (6.8\% column mention), they still assume pre-identified databases. No existing benchmark evaluates the full pipeline from colloquial ambiguous queries through autonomous dataset discovery to validated analysis and reporting.} This ensures each question can be answered end-to-end with verifiable public data, supporting reproducibility and real-world relevance.

\begin{table}[htbp]
\centering
\caption{Benchmark Query Domains and Representative Examples}
\label{tab:benchmark_domains}
\begin{tabularx}{\textwidth}{lX}
\toprule
Domain & Example Query \\
\midrule
Health Epidemiology & ``How prevalent is adult high blood pressure in the United States?'' \\
Environmental Monitoring & ``Which NYC neighborhood has the worst nitrogen dioxide air pollution?'' \\
COVID-19 Analysis & ``How do COVID-19 death rates compare between unvaccinated and vaccinated people in Chicago in 2022?'' \\
Transportation & ``Which area in Washington has the most electric cars?'' \\
Campaign Finance & ``How have total campaign contributions changed from 2020 to 2025 in Washington state?'' \\
Public Health & ``Which racial or ethnic group has had the highest cancer death rate in NYC since 2007?'' \\
\bottomrule
\end{tabularx}
\end{table}

\begin{table}[htbp]
\centering
\caption{Analytical Intricacy Stratification with Examples}
\label{tab:intricacy_levels}
\begin{tabularx}{\textwidth}{lX}
\toprule
Intricacy Level & Example Query \\
\midrule
Easy (Direct Aggregation) & ``Which state has the highest rate of obese adults?'' \\
Medium (Filtering and Grouping) & ``How many electric cars in Seattle are fully electric, not hybrids?'' \\
Hard (Trend Analysis) & ``Has the obesity rate in California gone up or down over the years?'' \\
\bottomrule
\end{tabularx}
\end{table}

Benchmark design determines what abilities we measure. Prior datasets like Spider and BIRD focus on semantic parsing accuracy, testing text-to-SQL skill but not dataset discovery or reporting \citep{yu2018spider,li2023bird}. Our benchmark includes the full pipeline from ambiguous natural language to evidence-backed reports. The stratified complexity captures reasoning differences among aggregation, filtering, and temporal analysis \citep{ribeiro2020checklist}. Domain diversity tests generalization, since models trained on narrow distributions often fail on unfamiliar topics.

\subsection{Models and Decoding Temperatures}
We evaluate five models from multiple sources: OpenAI GPT~OSS~120B, Meta Llama~3.3~70B Instruct, OpenAI GPT-4o Mini, X.AI Grok~3 Mini, and Google Gemini~2.5 Pro (Table~\ref{tab:model_config}). Differences in training and size make results less dependent on one model family. Each model’s decoding temperature $\tau$ is tuned on a held-out set to balance quality and consistency.

Temperature affects the balance between determinism and diversity \citep{holtzman2020curious}. Low $\tau$ favors consistent but repetitive text. High $\tau$ increases variation but can cause factual errors. We select $\tau$ to reduce output variance while keeping analytical completeness. Models like Llama, tuned for instruction following, often need higher $\tau$ to avoid rigid patterns, while RLHF-trained models may perform best at lower $\tau$ \citep{renze2024temperature,ouyang2022training}. The final values in Table~\ref{tab:model_config} reflect each model’s optimum setting.

\begin{table}[htbp]
\centering
\caption{Language Model Configuration for Evaluation. Parameter counts marked with $\dagger$ are not public.}
\label{tab:model_config}
\begin{tabularx}{\textwidth}{lXX}
\toprule
Model & Parameters & Temperature \\
\midrule
Google Gemini 2.5 Pro & Large$^\dagger$ & 1.0 \\
Meta Llama 3.3 70B Instruct & 70B & 0.9 \\
OpenAI GPT-4o Mini & Small$^\dagger$ & 1.0 \\
OpenAI GPT OSS 120B & 120B & 0.5 \\
X.AI Grok 3 Mini & Small$^\dagger$ & 1.0 \\
\bottomrule
\end{tabularx}
\end{table}

\subsection{Metrics and Mathematical Formulation}
Each system produces a report $R$ for query $q$. An independent judge scores factual consistency $F$, completeness $C$, relevance $V$, and coherence $H$ on a 1-10 scale using explicit rubrics:

\begin{itemize}
    \item \textbf{Factual Consistency (1-10):} Does the report maintain internal logic without contradictions? Do the data points match the analysis? Are claims supported by the evidence presented?
    \item \textbf{Completeness (1-10):} Does the report cover key aspects of a data analysis? Are all necessary components present for a thorough analysis?
    \item \textbf{Relevance (1-10):} Does the report directly answer the question? Is the content focused and on-topic?
    \item \textbf{Coherence (1-10):} Does the report have logical flow and readability? Is it well-structured and easy to follow?
\end{itemize}

The overall score is computed as the average:
\[
Q(R) = \tfrac{1}{4}(F(R) + C(R) + V(R) + H(R)).
\]
For ablations, we compare full outputs $R^{\text{full}}$ with ablated versions $R^{\text{abl}}$ using blinded pairwise judgments. The win rate for criterion $k$ is
\[
W_k = \frac{\text{\# queries where } R^{\text{full}} \text{ wins on } k}{50}.
\]
We report $W_k$ across all 50 benchmark queries.

\subsection{LLM-as-Judge Protocol and Related Practice}
A strong independent LLM acts as judge, scoring reports and deciding A/B preferences using a fixed rubric. This approach aligns with MT-Bench and Chatbot Arena, which show strong correlation between LLM and human ratings \citep{zheng2023judging}. We apply position randomization, length normalization, and explicit rubrics following AlpacaEval~2.0 and Prometheus \citep{alpacaeval2024,prometheus2024}. These steps reduce bias and improve reproducibility \citep{gu2024survey,chen2024bias}.

LLM judges are efficient for multi-dimensional quality evaluation. Human review of long-form text is costly and inconsistent. LLMs apply criteria uniformly and provide explanations for their scores \citep{zheng2023judging}. Our rubrics define concrete evidence—citations, numerical checks, logical order—rather than vague descriptors, improving reliability \citep{dubois2024length}. While not perfect human substitutes, LLM judges perform well on structural criteria like coherence and completeness.

For pairwise ablation comparisons, the judge evaluates both reports on four criteria with explicit instructions:

\begin{itemize}
    \item \textbf{Factual Consistency:} Which report maintains better internal logic without contradictions? Do the data points match the analysis better?
    \item \textbf{Completeness:} Which report better covers intent clarification, data sourcing, analysis steps, and conclusions?
    \item \textbf{Relevance:} Which report more directly answers the question without unnecessary fluff?
    \item \textbf{Coherence:} Which report has better logical flow and readability?
\end{itemize}

\textcolor{blue}{These qualitative comparisons are converted to quantitative metrics through structured scoring. For each query $q$ and criterion $k$, the judge assigns a binary outcome: win (1), loss (0), or tie (0.5). The criterion-specific win rate is computed as:
\[
W_k = \frac{1}{|\mathcal{Q}|}\sum_{q \in \mathcal{Q}} s_k(q),
\]
where $s_k(q) \in \{0, 0.5, 1\}$ represents the outcome for criterion $k$ on query $q$. Aggregating across all queries yields the percentage of wins for each evaluation dimension.}

The judge responds with structured JSON output indicating the winner for each criterion with explanation, plus an overall winner determination. This structured format ensures consistent evaluation across all comparisons and enables quantitative aggregation of win rates.

\subsection{Ablations: Testing the Multi-Agent Hypothesis}
We remove one agent at a time—intent clarification, data discovery, analysis, or reporting—while keeping the orchestrator to fill the gap. This isolates each agent’s contribution.

Ablation testing isolates component effects in complex systems \citep{meyes2019ablation}. It attributes performance changes to architecture, not confounding factors such as model size. We maintain comparable workflows, vary one component per test, and measure effect size through scores and pairwise preferences. Consistent effects across model families support general conclusions.

Across five models, the full system wins most pairwise comparisons: analysis-agent ablations lose 76-100\% and discovery-agent ablations lose 79.5-98\%. These results confirm that specialized agents are required for consistent performance across architectures and sizes.

\subsection{Comparative Protocol and Validity}
For each $q\in\mathcal{Q}$, we generate both full and ablated reports. Judges assign rubric scores and blinded preferences to compute $Q(R)$ and $W_k$. We reduce bias by randomizing order, allowing ties, and anchoring rubrics. Some residual bias may persist. The 50-query set is balanced and grounded in open-data portals, but expanding to more domains and formats would improve coverage. Despite this, consistent ablation outcomes across all models show that agent specialization improves open-data analysis and reporting beyond what scale alone achieves.

%% file: sections/results.tex
\section{Results}

Our findings confirm that \modelname performs strongly across all evaluated models and that its multi-agent structure offers measurable advantages.

\subsection{Benchmark Performance Analysis}

Table~\ref{tab:benchmark_results} shows the performance of all models across factual, completeness, relevance, and coherence criteria.

\begin{table}[htbp]
\centering
\caption{Benchmark Performance Results by Model and Quality Criteria}
\label{tab:benchmark_results}
\begin{tabularx}{\textwidth}{lXXXXX}
\toprule
Model & Factual & Complete & Relevant & Coherent & Overall \\
\midrule
Google Gemini 2.5 Pro & 6.8 & 6.3 & 7.4 & 8.2 & 7.2 \\
Meta Llama 3.3 70B Instruct & 4.7 & 3.3 & 4.1 & 6.7 & 4.7 \\
OpenAI GPT-4o Mini & 5.5 & 5.6 & 7.9 & 8.1 & 6.8 \\
OpenAI GPT OSS 120B & 7.1 & 7.9 & 8.5 & 9.2 & 8.2 \\
X.AI Grok 3 Mini & 5.1 & 4.8 & 5.2 & 7.8 & 5.8 \\
\bottomrule
\end{tabularx}
\end{table}

OpenAI GPT OSS 120B ranks highest with an overall score of 8.2, showing strong coherence (9.2) and relevance (8.5). This indicates consistent and well-structured analytical outputs. Google Gemini 2.5 Pro follows with balanced results across all criteria (7.2 overall).

OpenAI GPT-4o Mini performs competitively at 6.8, despite being smaller in scale. Its strong relevance (7.9) and coherence (8.1) scores show that compact models can still excel in structured reasoning tasks when properly supported by agent architecture.

Meta Llama 3.3 70B performs poorly (4.7), demonstrating that parameter count alone does not predict success in multi-agent analytical settings. The low scores confirm that training quality and architecture matter more than model size for complex analytical tasks.

Completeness remains the hardest criterion, with wide score variation (3.3–7.9). This pattern shows that generating thorough reports across multiple data dimensions is the most demanding aspect of the task.

\subsection{Performance Across Question Difficulty}

Table~\ref{tab:difficulty_performance} presents results by question difficulty that we manually assigned during the benchmark dataset creation process, based on perceived complexity of the data analysis task.

\begin{table}[htbp]
\centering
\caption{Benchmark Performance by Question Difficulty}
\label{tab:difficulty_performance}
\begin{tabularx}{\textwidth}{lXXXX}
\toprule
Model & Easy & Medium & Hard & Overall \\
\midrule
Google Gemini 2.5 Pro & 7.3 & 7.4 & 6.5 & 7.2 \\
Meta Llama 3.3 70B Instruct & 4.5 & 5.3 & 3.9 & 4.7 \\
OpenAI GPT-4o Mini & 7.1 & 7.1 & 5.8 & 6.8 \\
OpenAI GPT OSS 120B & 8.1 & 8.2 & 8.2 & 8.2 \\
X.AI Grok 3 Mini & 5.2 & 6.4 & 5.1 & 5.8 \\
\bottomrule
\end{tabularx}
\end{table}

OpenAI GPT OSS 120B maintains stable results across all difficulty levels (8.1–8.2), indicating consistent reasoning performance. Gemini 2.5 Pro and GPT-4o Mini both show mild declines on hard questions (0.8–1.3 points), though they handle easier queries well.

Meta Llama 3.3 70B and X.AI Grok 3 Mini show modest variation across difficulty levels, with no consistent degradation pattern. This suggests that difficulty alone does not fully explain performance variation—other factors such as query structure and domain may also play important roles.

\subsection{Criterion-Specific Ablation Effects}

Table~\ref{tab:criterion_ablation_results} summarizes how removing agents affects quality dimensions. Each agent influences distinct aspects of performance, showing complementary rather than redundant roles.

\begin{table}[htbp]
    \centering
    \caption{Full System Win Rates by Criterion, Model, and Ablation Type}
    \label{tab:criterion_ablation_results}
    \begin{tabularx}{\textwidth}{llXXXX}
    \toprule
    Model & Ablation & Factual & Complete & Relevant & Coherent \\
    \midrule
    Google Gemini 2.5 Pro & No Analysis & 71.4\% & 79.6\% & 69.4\% & 71.4\% \\
     & No Discovery & 69.4\% & 83.7\% & 83.7\% & 79.6\% \\
     & No Report & 53.1\% & 63.3\% & 69.4\% & 55.1\% \\
     & No Intent & 61.2\% & 73.5\% & 69.4\% & 73.5\% \\
    \midrule
    Meta Llama 3.3 70B Instruct & No Analysis & 76.7\% & 100.0\% & 79.1\% & 97.7\% \\
     & No Discovery & 69.2\% & 82.1\% & 61.5\% & 87.2\% \\
     & No Report & 57.8\% & 93.3\% & 73.3\% & 93.3\% \\
     & No Intent & 67.3\% & 89.8\% & 79.6\% & 91.8\% \\
    \midrule
    OpenAI GPT-4o Mini & No Analysis & 42.0\% & 88.0\% & 76.0\% & 76.0\% \\
     & No Discovery & 64.0\% & 96.0\% & 92.0\% & 88.0\% \\
     & No Report & 44.0\% & 68.0\% & 62.0\% & 46.0\% \\
     & No Intent & 63.3\% & 73.5\% & 63.3\% & 69.4\% \\
    \midrule
    OpenAI GPT OSS 120B & No Analysis & 96.0\% & 96.0\% & 80.0\% & 92.0\% \\
     & No Discovery & 94.0\% & 100.0\% & 100.0\% & 100.0\% \\
     & No Report & 96.0\% & 100.0\% & 100.0\% & 100.0\% \\
     & No Intent & 98.0\% & 98.0\% & 98.0\% & 98.0\% \\
    \midrule
    X.AI Grok 3 Mini & No Analysis & 82.0\% & 100.0\% & 98.0\% & 100.0\% \\
     & No Discovery & 88.0\% & 100.0\% & 96.0\% & 98.0\% \\
     & No Intent & 90.0\% & 100.0\% & 96.0\% & 98.0\% \\
    \bottomrule
    \end{tabularx}
    \end{table}

\textcolor{blue}{Win rates indicate the percentage of queries where the full multi-agent system outperforms the ablated version missing one agent, with higher percentages showing stronger agent contribution.}

Removing the analysis agent reduces factual accuracy most sharply, with win rates from 42\% (GPT-4o Mini) to 96\% (GPT OSS 120B). This wide variation shows that models differ in how they rely on structured reasoning—some need agents for factual verification, others for structural organization. The analysis agent consistently supports completeness (88–100\% win rates), confirming its importance for full analytical coverage regardless of model strength.

The discovery agent achieves near-perfect completeness gains (96–100\% win rates) and strong relevance effects (61–100\% win rates). This pattern shows that proper dataset discovery improves both scope and precision. The dual effect confirms that discovery provides the right information and keeps analysis on topic.

Report generation has balanced effects across all criteria, with win rates from 44–96\% (factual), 63–100\% (completeness), 62–100\% (relevance), and 46–100\% (coherence). This uniformity shows the agent's integrative role. The large performance gap between models (GPT OSS 120B: 96–100\% vs GPT-4o Mini: 44–68\%) indicates that smaller models depend more on this agent for coherent synthesis.

Intent clarification strongly improves completeness (73–100\% win rates) and shows moderate effects elsewhere (61–98\%). This pattern supports the idea that clear query definition benefits all downstream processing without targeting any single quality dimension.

\subsection{Model-Specific Agent Dependencies}

Agent dependencies differ sharply across models, showing that optimal architecture depends on base model characteristics.

OpenAI GPT OSS 120B benefits uniformly from all agents (92–100\% win rates), showing broad synergy without overreliance on any one component. No single agent is disproportionately important. Even this strong base model (8.2/10) cannot compensate for missing agents through its own capabilities.

In contrast, Meta Llama 3.3 70B shows selective dependence on analysis and report agents, with 76–100\% win rates when analysis is removed and 57–93\% for report generation. The particularly strong reliance on completeness (100\% analysis, 93\% report) shows the model struggles with comprehensive output generation and needs substantial architectural support.

OpenAI GPT-4o Mini depends most on the analysis agent for factual grounding, showing only 42\% wins on factual consistency when this agent is removed—the lowest across all models. This reveals a specific weakness in maintaining factual accuracy during complex analytical tasks. The analysis agent provides critical verification that the base model cannot perform independently, showing that smaller models can achieve competitive overall performance through strategic agent support that compensates for specific weaknesses.

X.AI Grok 3 Mini performs well with most agents (82–100\% win rates) and achieves perfect completeness scores (100\%) when other agents are removed. However, the notable absence of successful report generation ablation data shows the model has fundamental difficulty with end-to-end report synthesis, requiring the specialized report agent for basic functionality rather than just quality improvement.

\subsection{Design Principles for Agentic AI Systems}

Our results reveal five fundamental principles about multi-agent architectures that extend beyond this specific application. Each principle is supported by quantitative evidence and provides actionable guidance for system designers.

\subsubsection{Specialization Provides Value Independent of Model Strength}

Multi-agent architecture improves performance regardless of base model capability, with benefits orthogonal to model scale. Comparing baseline model performance (4.7–8.2/10) with average agent win rates (72–98\%) shows no inverse correlation. Even the strongest model (GPT OSS 120B, 8.2/10 baseline) shows 97.5\% average agent wins across all ablations. \textcolor{blue}{This finding shows that agent-based coordination addresses workflow challenges—attention management, validation, information flow—that model scale alone cannot solve. Stronger models benefit from specialization for coordination and completeness rather than compensating for capability gaps.}

\textcolor{blue}{Organizations should deploy multi-agent architectures even with frontier models like GPT-4, budgeting for agent infrastructure from project inception rather than treating it as an optional enhancement for weaker models. The evidence contradicts the assumption that upgrading to a stronger base model eliminates the need for specialization—even the most capable models require architectural support for complex workflows.}

\subsubsection{Agent Utility Varies by Function: Universal vs Conditional}

Agents differ in their universality—some provide consistent value across models (universal agents), while others show model-dependent effectiveness (conditional agents). The discovery agent shows low variance in win rates (std dev 12.4\%), providing uniform benefit across all models. In contrast, the report generation agent shows high variance (std dev 20.5\%), dramatically helping some models while barely affecting others. \textcolor{blue}{This pattern shows that data discovery is a universal bottleneck all models face, while report synthesis depends heavily on base model generation capabilities. Universal agents (discovery, analysis) address architectural challenges independent of model choice, while conditional agents (report, intent) compensate for model-specific weaknesses.}

\textcolor{blue}{System designers should prioritize universal agents (discovery, analysis) for all deployments. For conditional agents (report, intent), conduct model profiling on 20-30 representative queries before deployment to measure effectiveness. In resource-constrained environments, deploy agents in priority order: first discovery, then analysis, and conditional agents only if profiling shows greater than 60\% win rates. Multi-model systems should maintain separate agent configurations for each model rather than applying uniform architecture, as the same conditional agent may be critical for one model but redundant for another.}

\subsubsection{Agents Mitigate Distinct Failure Types: Non-Redundant Specialization}

Each agent addresses a distinct failure mode rather than redundantly solving the same problem, with failures manifesting differently based on which agent is removed. Analysis of judge explanations reveals two failure categories. Removing discovery or analysis agents causes complete task failure (243–280 instances) where the system produces no meaningful output. Removing report or intent agents causes degraded quality—reduced detail or logical inconsistencies—while still generating output. \textcolor{blue}{This hierarchy shows that discovery and analysis form critical infrastructure (their absence causes catastrophic failure), while report and intent provide quality enhancement (their absence causes degradation). The distinct failure signatures show that agents are non-redundant—each addresses a specific failure mode that others cannot compensate for.}

\textcolor{blue}{These failure patterns provide actionable diagnostics when debugging multi-agent systems. Empty or nonsensical outputs point to issues with discovery or analysis agents. Outputs using wrong data sources point to discovery agent problems, while calculation errors point to analysis agent issues. Outputs lacking structure point to report agent problems, and outputs addressing the wrong question point to intent agent issues. For staged deployment, establish discovery and analysis infrastructure first as the critical path, then add report and intent agents for quality enhancement once the core functionality is stable.}

\subsubsection{Architectural Benefits Persist Across Task Complexity}

Agent-based architectures provide consistent value across task complexity levels, addressing structural workflow challenges rather than just complex reasoning. Agent win rates remain stable across question difficulty with no systematic variation: analysis (86–92\%), discovery (84–94\%), report (71–79\%), and intent (81–87\%). If agents primarily helped with difficult reasoning, we would expect dramatically higher win rates on hard questions. \textcolor{blue}{The stable pattern shows that agents address structural challenges—task decomposition, information flow, output integration—that affect all questions regardless of inherent complexity. This shows that agent benefits stem from workflow management rather than enhanced reasoning capability, guiding future agent design toward coordination mechanisms rather than pure intelligence augmentation.}

\textcolor{blue}{Organizations should budget for agent infrastructure across all query types rather than treating agents as optional assistance for complex tasks. Dynamic agent activation based on estimated query complexity hurts performance—even simple questions benefit from the full architecture. If cost reduction is necessary, achieve it through model selection strategies where smaller base models with agents can outperform larger models without agents, rather than through selective agent removal that degrades quality uniformly across all complexity levels.}

\subsubsection{Model-Agent Fit Determines Effectiveness: Architecture Must Adapt to Model}

Optimal agent architecture depends on model-specific strengths and weaknesses, with the same agent providing drastically different value across models. The analysis agent shows 42–96\% win rate variance across models. GPT-4o Mini requires the analysis agent primarily for factual grounding (42\% win rate on factual consistency), while GPT OSS 120B benefits uniformly (92–96\% across all criteria). Weaker models need agents for basic functionality (correctness), while stronger models benefit from agents for coordination and completeness. \textcolor{blue}{This variance shows that no single agent configuration is optimal across all models—model-specific weaknesses determine which agents provide the most value, and applying uniform agent architectures across different models wastes resources on low-value agents while potentially under-investing in high-value ones.}

\textcolor{blue}{Before production deployment, system designers should conduct model profiling: evaluate the base model on 20-50 representative queries without agents, categorize weaknesses by failure type (factual errors suggest analysis agent need, missing data suggests discovery agent need, poor structure suggests report agent need, wrong focus suggests intent agent need), run targeted ablations to measure each agent's win rate, then activate agents showing greater than 60\% win rates while deferring agents below 40\%. For production systems serving multiple models, maintain model-specific agent configurations and implement adaptive routing that activates agents based on runtime model capability assessment for specific query types, balancing quality maintenance with inference cost optimization.}

%% file: sections/conclusion.tex
\section{Conclusion}

We derived five design principles for multi-agent LLM systems through systematic evaluation of \modelname, a framework for open data analysis. These principles establish when and why specialization is necessary for complex analytical workflows: (1) multi-agent architectures provide value independent of model strength, (2) agents divide into universal and conditional categories requiring different deployment strategies, (3) agents mitigate distinct failure modes rather than redundantly solving the same problems, (4) architectural benefits persist across task complexity indicating workflow management value, and (5) optimal architecture must adapt to model-specific strengths and weaknesses.

Through ablation studies across five models and 50 queries, we demonstrate that these principles hold regardless of base model capability. Even the strongest model shows 97.5\% agent win rates, confirming benefits orthogonal to model scale. The wide variance in agent effectiveness across models (42–96\%) provides actionable guidance for system designers: profile models before deployment, prioritize universal agents (discovery, analysis) for all systems, and make conditional agents (report, intent) dependent on identified model weaknesses. These principles extend beyond open data analysis to inform design of multi-agent systems for complex, multi-stage workflows.

Beyond establishing design principles, \modelname demonstrates practical impact by expanding access to public datasets through natural language interfaces. By reducing technical barriers for journalists, policymakers, and the public, it supports transparent, evidence-based decisions. Future research should validate these principles across additional domains, develop adaptive agent routing based on runtime capability assessment, and explore how coordination mechanisms can be optimized for emerging model architectures that balance effectiveness with computational efficiency.